\documentclass[letterpaper, 10 pt, conference]{ieeeconf}  %

\IEEEoverridecommandlockouts                              %

 \usepackage[nocompress]{cite}
 \usepackage{amsmath}
 \usepackage[utf8]{inputenc}
 \usepackage{graphicx}
 \usepackage{lipsum}
 \usepackage{makecell}
 \usepackage{threeparttable}
 \usepackage{cleveref}
 \usepackage{calc}
 \usepackage{tabularx}
 \usepackage{float}
 \usepackage{microtype}
\usepackage{amsmath} %
\usepackage{amssymb}  %

\DeclareMathOperator*{\argmax}{arg\,max}

\title{\LARGE \bf
VPRTempo: A Fast Temporally Encoded Spiking Neural Network for Visual Place Recognition}

\author{Adam D Hines \quad\qquad Peter G Stratton \quad\qquad Michael Milford \quad\qquad Tobias Fischer%
\thanks{The authors are with the QUT Centre for Robotics, School of Electrical Engineering and Robotics, Queensland University of Technology, Brisbane, QLD 4000, Australia. Email:
        {\tt adam.hines@qut.edu.au}}%
\thanks{This work received funding from Intel Labs to TF and MM, AUSMURIB000001 associated with ONR MURI grant N00014-19-1-2571 and an ARC Laureate Fellowship FL210100156 to MM, and the Commonwealth of Australia as represented by the Defence Science and Technology Group of the Department of Defence to PGS. The authors acknowledge continued support from the Queensland University of Technology (QUT) through the Centre for Robotics.}
}
\usepackage{eso-pic}
\usepackage{url}
\AddToShipoutPicture*{%
     \AtTextUpperLeft{%
         \put(-3.5,10){
           \begin{minipage}{\textwidth}
              \scriptsize
              \MakeUppercase{IEEE International Conference on Robotics and Automation (ICRA) 2024.\\
              Preprint version; Final version will be available at}
              \url{https://ieeexplore.ieee.org/}
           \end{minipage}}%
     }%
}

\begin{document}

\maketitle
\thispagestyle{empty}
\pagestyle{empty}

\begin{abstract}

Spiking Neural Networks (SNNs) are at the forefront of neuromorphic computing thanks to their potential energy-efficiency, low latencies, and capacity for continual learning. While these capabilities are well suited for robotics tasks, SNNs have seen limited  adaptation in this field thus far. This work introduces a SNN for Visual Place Recognition (VPR) that is both trainable within minutes and queryable in milliseconds, making it well suited for deployment on compute-constrained robotic systems. Our proposed system, VPRTempo, overcomes slow training and inference times using an abstracted SNN that trades biological realism for efficiency. VPRTempo employs a temporal code that determines the \emph{timing} of a \emph{single} spike based on a pixel's intensity, as opposed to prior SNNs relying on rate coding that determined the \emph{number} of spikes; improving spike efficiency by over 100\%. VPRTempo is trained using Spike-Timing Dependent Plasticity and a supervised delta learning rule enforcing that each output spiking neuron responds to just a single place. We evaluate our system on the Nordland and Oxford RobotCar benchmark localization datasets, which include up to 27k places. We found that VPRTempo's accuracy is comparable to prior SNNs and the popular NetVLAD place recognition algorithm, while being several orders of magnitude faster and suitable for real-time deployment -- with inference speeds over 50 Hz on CPU. VPRTempo could be integrated as a loop closure component for online SLAM on resource-constrained systems such as space and underwater robots.
\end{abstract}

\section{Introduction}

Spiking neural networks (SNNs) are computational tools that model how neurons in the brain send and receive information \cite{brainsci12070863}. SNNs have attracted significant interest due to their energy efficiency, low-latency data processing, especially when deployed on neuromorphic hardware such as Intel's Loihi or SynSense's Speck \cite{6330636,orchard2021efficient}. However, modeling the complexity of biological neurons limits the computational efficiency of SNNs, especially where resource-constrained systems with real-time applications are concerned. Using a SNN system that trades biological realism for overall system efficiency would be well suited for a variety of robotics tasks \cite{scirobcere,vitale2021eventdriven,snnrobothand,snnroboarm}.

    \begin{figure*}
    \includegraphics[width=\textwidth]{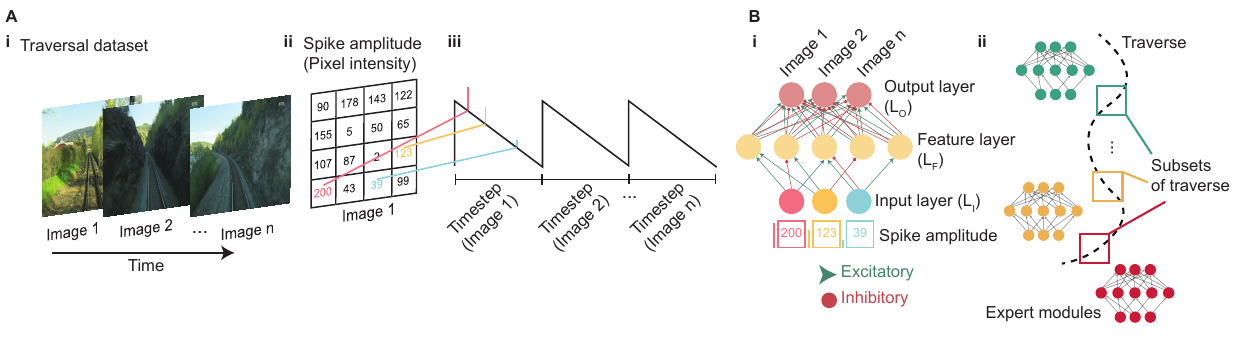}
    \caption{\textbf{A-i} Image sequences from standard VPR datasets (Nordland, Oxford RobotCar) are filtered and processed to be converted into \textbf{A-ii} spikes where the pixel intensity determines amplitude. \textbf{A-iii} In order to temporally encode spikes to an abstracted theta oscillation, amplitudes determine the spike timing during a timestep. \textbf{B-i} Once spikes have been generated, they are passed into a SNN with 3 layers; an input, feature layer, and a one-hot encoded output layer where each output neuron represents one place. \textbf{B-ii} In order to scale the system for large datasets, we train individual expert module SNNs of up to 1000 places from subsets of an entire reference dataset.}
    \vspace*{-0.3cm}
    \label{fig:schema}
    \end{figure*}

One such task that could benefit from fast and efficient SNNs is visual place recognition (VPR), where incoming query images are matched to a potentially very large reference database, with a range of applications in robot localization and navigation \cite{hussaini2022ensembles,hussaini2022spiking,SchubertVisual,Arandjelovic16,doi:10.1126/scirobotics.abm6996,7298790,leyvavallina2021gcl,alibey2023mixvpr,sarlin2018leveraging}, including as a loop closure component in Simultaneous Localization and Mapping (SLAM) \cite{slamessentialalg,loopclose,SchubertVisual,realtimecnn,Wang_2020}. However, VPR remains challenging as query images often have significant appearance changes when compared to the reference images, with factors such as time of day, seasonal changes, and weather contributing to these differences \cite{Zaffar2021,9336674,7339473}. SNNs represent a unique way to perform VPR for their low-latency and energy efficiency, especially where this is an important consideration for the overall design of a robot \cite{roboenergydesign}. The flexible and adaptive learning that SNNs take advantage of make them ideal for VPR, as network connections and weights will learn unique place features from reference datasets.

SNNs employ a variety of encoding mechanisms, with the traditional rate encoding suffering from substantial computational costs due to its reliance on spike counts \cite{eshraghian2023training,wenzhesnn,firstspikesspatialfingertips}. Our work instead adopts a temporal encoding strategy inspired by central information storage techniques used by the brain, reducing the overall content the network processes during learning \cite{ratetotemporal,stratton2022making}. This takes a different approach to what information each spike is representing and transmitting when compared to previous studies like those by Hussaini et al. \cite{hussaini2022ensembles,hussaini2022spiking}. We leverage a simplified SNN framework \cite{stratton2022making}, modular networks \cite{doi:10.1142/S0129065799000125,HAPPEL1994985,6797059}, and one-hot encoding for supervised training to substantially decrease training and inference times. Compared with previous systems, our SNN achieves query speeds exceeding 50 Hz on CPU hardware for large reference databases ($>$27k places) and even higher inference speeds when deployed on GPUs (as high as 500 Hz), indicating its potential usefulness for real-time applications \cite{hussaini2022ensembles,6224623,hausler2021patchnetvlad,Arandjelovic16,leyvavallina2021gcl,Liu_2019_ICCV,Nowicki2017,doi:10.1126/scirobotics.abm6996}.

\noindent Specifically, the key contributions of this work are:
    \begin{enumerate}
        \item 
            We present VPRTempo, a novel SNN system for VPR that, to the best of our knowledge, for the first time encodes place information via a temporal spiking code (Figure \ref{fig:schema}), significantly increasing the information content of each spike.
        \item    
            We significantly lower the training time of spiking networks to under an hour \emph{and} increase query speeds to real-time capability on both CPUs and GPUs, with the potential to represent tens of thousands of places even in resource-limited compute scenarios.
        \item 
            We demonstrate that our lightweight and highly compute efficient system results in comparable performance to popular place recognition systems like NetVLAD \cite{Arandjelovic16} on the Nordland \cite{olid2018single} and Oxford \mbox{RobotCar} \cite{oxfordrobot} benchmark datasets.
    \end{enumerate}

\noindent To foster future research and development, we have made the code available at:\medskip

\texttt{https://github.com/QVPR/VPRTempo}

\section{Related Works}

In this section, we review previous works about VPR (\Cref{subsec:VPR}), how robotics have used and deployed SNNs (\Cref{subsec:RoboticsSNNs}), using SNNs to perform VPR tasks (\Cref{subsec:SNNsVPR}), and using temporal encoding in SNNs (\Cref{subsec:temporalsnn}).

\subsection{Visual place recognition}
\label{subsec:VPR}
Visual place recognition (VPR) is an important task in robotic navigation and localization for an accurate, and ideally comprehensive, representation of their environment and surroundings. The basic principle of VPR is to match single or multiple query image sets to previously seen or traversed locations in a reference database \cite{SchubertVisual,9336674,7339473,Zaffar2021}. This remains challenging for numerous reasons, including but not limited to: changes in viewpoint, time of day, or severe weather fluctuations \cite{Garg_2021,9197044}.

Several systems have been deployed to tackle this problem, ranging from simple approaches such as Sum of Absolute Differences (SAD)~\cite{6224623}, over handcrafted systems like \mbox{DenseVLAD} \cite{7298790} to learned approaches like NetVLAD \cite{6224623,Arandjelovic16} and many others including \cite{leyvavallina2021gcl,alibey2023mixvpr,doi:10.1126/scirobotics.abm6996}. These systems are commonly used in localization tasks such as loop closure in Simultaneous Localization And Mapping (SLAM) \cite{Tsintotas_2022}. Although these approaches have shown to be highly effective at accurate place recognition, they require computational and time intensive training regimes and learn generalized feature extraction, rather than actually learning the specific location. 

In this work, we are interested in alternative spiking neural network solutions to approach place recognition, which have desirable characteristics such as being able to have online adaptation and energy efficiency, especially when deployed on specialized neuromorphic hardware \cite{orchard2021efficient,6330636,Pei2019,7229264}.

\subsection{SNNs in robotics}
\label{subsec:RoboticsSNNs}
Thanks to these characteristics, SNNs have been deployed to perform a variety of robotics tasks, including in the physical interaction domain for robotic limb control and manipulation \cite{snnrobothand}, scene understanding \cite{10.3389/fnins.2020.00551}, and object tracking \cite{9486868}. Of significant interest is the potential to combine neuromorphic algorithms and hardware for these tasks \cite{vitale2021eventdriven}, since this has the capacity to decrease compute time and conserve energy consumption where this is a concern (e.g.~battery powered autonomous vehicles). However, it is important to note that the deployment and use cases of SNNs in robotics is limited and has been mostly constrained to simulations or indoor/small-scale environments \cite{8594228,hussaini2022ensembles}.

\subsection{Spiking neural networks for robot localization}
\label{subsec:SNNsVPR}
Neuromorphic computing involves the development of hardware, algorithms, and sensors inspired by neuroscience to solve a variety tasks \cite{doi:10.1126/scirobotics.abm6996,9340948,8967864,vitale2021eventdriven,zora149361}. Clearly, the brain is robustly capable of place recognition in a variety of contexts, providing a solid rationale to explore neuromorphic systems for VPR related challenges. A recent study exhibited prolonged training and querying times in an SNN, due to the reliance on the low spike efficiency of rate encoding and modeling complex neuronal dynamics \cite{hussaini2022ensembles}. Robotic localization tasks have also employed SNNs to perform SLAM \cite{6418493,8967864}, encode a contextual cue driven multi-modal hybrid sensory system for place recognition \cite{doi:10.1126/scirobotics.abm6996}, and navigation \cite{LIU2021104362,Giraldo2023}. Our approach mitigates traditional barriers to deploying SNNs for real-time compute scenarios, enhancing the learning speed substantially by simplifying network processes and incorporating temporal spike encoding.

\subsection{Using temporal encoding in spiking networks}
\label{subsec:temporalsnn}
Temporal encoding has, to date, not been utilized for VPR tasks but is a common method used in SNNs \cite{10.3389/fnins.2021.712667,Auge2021,9533837,ratetotemporal}. Sometimes referred to as a latency code, rather than using the number of spikes to encode information, the timing of the spike is taken into consideration. This is particularly an important concept when designing a system that updates weights using spike timing dependent plasticity (STDP) \cite{Bi10464}, as the temporal information of a pre-synaptic spiking neuron can help determine which post-synaptic neurons to connect to.

Various temporal coding strategies have been used for image classification where it achieved very high accuracy \cite{Chen_Qu_Zhang_Wang_2021,9533837}. One way to achieve temporal coding is to define the pixel intensity of an image not by the number of spikes the system propagates, but the timing of a single spike in a time-step \cite{stratton2022making}. Another method to achieve a similar outcome is to model the oscillatory activity of the brain to modulate when spikes occur, relative to the phase of a constant, periodic shift in a neurons internal voltage \cite{orchard2021efficient,IZHIKEVICH2001883,alkhamissi2021deep}.

\section{Methodology}
This work naturally expands previously published works on SNNs for VPR. Hussaini et al.~\cite{hussaini2022ensembles} have introduced a scalable means of representing places, whereby independent networks represent geographically distinct regions of the traverse. However, as reviewed in Section \ref{subsec:SNNsVPR}, their network used rate coding with complex neuronal dynamics, which resulted in significant training times and non-real-time deployment. Here, by leveraging the temporal coding in conjunction with spike forcing proposed in BliTNet \cite{stratton2022making}, we overcome these limitations and demonstrate that BliTNet, combined with modularity, results in a high-performing VPR solution. We have completely re-implemented BliTNet in PyTorch \cite{NEURIPS2019_9015}, optimizing for computational efficiency and fast training times, thereby making it well-suited for deployment on compute-bound systems.%

This section is structured as follows: We first briefly introduce the underlying BliTNet in \Cref{sec:temporalcoding}. This is followed by the modular organization of networks in \Cref{sec:modularity}. We then describe our novel efficient parallelizable training and inference procedure in \Cref{sec:efficientimplementation}. 
\subsection{Temporal coding for visual place recognition}
\label{sec:temporalcoding}

\textbf{Network architecture:} Each network consists of 3 layers. The input layer $L_I$ has as many neurons as the number of pixels in the input image, i.e.~$W\cdot H$ neurons with $W$ and $H$ being the width and height of the input image, respectively. Neurons in $L_I$ are sparsely connected to a feature layer $L_F$. The $L_F$ layer is fully connected to the output layer $L_O$, which is one-hot encoded, such that each neuron $n_i \in L_O$ corresponds to one geographically distinct place from a set of reference images $\mathcal{D}=\{p_1, p_2, ..., p_N\}$. The number of output neurons $|L_O|$ matches the number of distinct places $N$. For training that uses multiple traversals over different days/times, the same output neuron is used to encode that place so the network can learn changes in environments. 

\textbf{Network connections:} Connections from $L_I \rightarrow L_F$ are sparse and determined by excitatory and inhibitory connection probabilities $P_{exc}$ and $P_{inh}$ respectively. Connections from $L_F \rightarrow L_O$ are fully connected, such that every neuron in $L_O$ receives both excitatory and inhibitory input. Excitatory and inhibitory connections are individually defined since inhibitory weights undergo additional normalization (refer to the Homeostasis subsection below).

\textbf{Neuron dynamics:} The neuron state $x^n_j$ for neuron $j$ in layer $n$ evolves in the following way:
    \begin{align}
    \begin{gathered}
        x^n_j = \sum^N_{m=1}\sum_ix^m_i(t)(W^{+nm}_{ji} - W^{-nm}_{ji}) + C - \theta^n_j,
    \label{eq:netstate}
    \end{gathered}
    \end{align}
where $x^m_i$ is input neuron $i$ in layer $m$, $\theta \in [0, \theta_{max}]$ is the neuron firing threshold, $C$ is a constant input, and $W^{+nm}_{ji}$ and $W^{-nm}_{ji}$ are the excitatory and inhibitory weights, respectively. $x^n_j$ is clipped in the range $[0, 1]$ to prevent spike vanishing or exploding, respectively.

\textbf{Weight updates and learning rules:} All weights between connected neurons are updated using spike timing dependent plasticity (STDP) rule \cite{Bi10464,stratton2022making}. STDP strengthens a connection between two neurons when the pre-synaptic cell fires before the post-synaptic, and vice versa. As the spike amplitude represents the timing in an abstracted theta oscillation or time-step, we can determine whether a pre- or post-synaptic neuron fired first using:

    \begin{align}
    \begin{gathered}
        \Delta W^{nm}_{ji}(t) =  \\  
        \frac{\eta_\text{STDP}(t)}{f^n_j} \cdot \Big[\Theta\big(x^m_i(t-1)\big)\cdot \Theta\big(x^n_j(t)\big)\cdot \big(0.5-x^n_j(t)\big)\Big],
    \label{eq:weightupdate}
    \end{gathered}
    \end{align}
    where $W^{nm}_{ij}$ refers to both positive and negative weights, $\eta_\text{STDP}$ is the STDP learning rate, $f_j^n$ is the target firing rate of neuron $j$ in layer $n$, $\Theta(\cdot)$ is the Heaviside step function, $x^m_i$ and $x^n_j$ are the neuron states of pre- or post-synaptic neurons respectively, and $t$ is the timestep. Effectively, a pre-synaptic neuron firing before a post-synaptic one will undergo a positive weight shift (synaptic potentiation). The inverse of this scenario will result in a negative weight shift (synaptic depression). To ensure that the weights persist as positive or negative connections, they cannot change sign during training. Therefore, if a sign change is detected then that connection will be reset to be $\epsilon = 10^{\pm 6}$, whereby the sign of $\epsilon$ matches the sign prior to the sign change.

The $\eta_\text{STDP}$ is initialized to $\eta^\text{init}_\text{STDP}$ and annealed during training according to:

    \begin{align}
    \begin{gathered}
        \eta_\text{STDP}(t) = \eta^\text{init}_\text{STDP}(1 - \tfrac{t}{T})^2,
    \label{eq:STDP}
    \end{gathered}
    \end{align}
where $t$ is the current time step and $T$ is the total number of training iterations.

\textbf{Homeostasis:} Inhibitory connections undergo additional normalization to balance and control excitatory connections. Whenever the sum input to a post-synaptic neuron is positive, the negative weights increase slightly. The inverse occurs when the net input is negative to post-synaptic neurons. Both cases are implemented by:

    \begin{align}
    \begin{gathered}
        \hat{W}^{-nm}_{ji}(t) \leftarrow W^{-nm}_{ji}(t)\Big(1-\eta_\text{STDP}(t)\cdot \Theta\big(\sum_i x^m_i(t)\big)\Big).
    \label{eq:homeostasis}
    \end{gathered}
    \end{align}

When the network is first initialized, neurons have randomly generated firing thresholds which may never be crossed depending on the input. To prevent neurons from being consistently inactive, an Intrinsic Threshold Plasticity (ITP) learning rate $\eta_\text{ITP}$ is used to adjust the spiking threshold to an ideal value for post-synaptic neurons:

    \begin{align}
    \begin{gathered}
        \Delta\theta^n_j(t) = \eta_\text{ITP}(t)\cdot\Big(\Theta\big(x^n_j(t)\big)-f^n_j\Big).
    \label{eq:theta}
    \end{gathered}
    \end{align}

If a threshold value goes negative, it is reset to 0. $\eta_\text{ITP}$ is initially set to $\eta^\text{init}_\text{ITP}$ and annealed as per Equation \ref{eq:STDP}.

\textbf{Spike forcing:} Spike forcing \cite{stratton2022making} was used in the output layer $L_O$ to create a supervised readout of the result that is represented in the feature layer $L_F$, similar to the delta learning rule \cite{Rumelhart1986}. For each learned place $p_i$, the assigned neuron $n_i$ in the output layer $L_O$ (refer to \ref{sec:temporalcoding} subsection Network Architecture) was forced to spike with an amplitude of $x_\text{force}=0.5$.

The network uses the differences of the calculated spikes from $L_F \rightarrow L_O$ to the $x_\text{force}$ to encourage the amplitudes to match. STDP learning rules strengthen connections between pre-synaptic neurons that cause output spikes and weakens those that do not:

    \begin{align}
    \begin{gathered}
        \Delta W^{nm}_{ji}(t) = \\
        \eta_\text{STDP}(t)/f^n_j . [x^m_i(t-1)(x_\text{force j}^n(t) - x^n_j(t)].
    \label{eq:spikeforce}
    \end{gathered}
    \end{align}

\subsection{Modular place representation for temporal codes}
\label{sec:modularity}

The brain encodes scene memory with place cells into sparse circuitry using engrams, individual circuits of neurons that activate in response to external cues for recall \cite{10.3389/fnbeh.2020.632019}. To mimic the concept of an engram we train multiple expert networks, which provides several advantages: 1) Smaller networks train faster and are more accurate, 2) the heterogeneity of initial network seeds is uniformly random, thus a query image $q$ trained in network $N_i$ is unlikely to activate output neurons with a high enough amplitude in other networks to generate a false positive match, and 3) similar to previous work \cite{hussaini2022ensembles}, modularizing improves system scalability to learn more places. 

Choosing the number of expert modules to employ depends on the dataset and number of places being encoded. Accuracy is influenced by $T$ and learning rate annealment therefore too few or too many places will affect the systems learning capability.

To that end, we employ a system that can simultaneously train non-overlapping subsets of images into separate networks $N_i$ and then test query images $q$ across all networks at once \cite{hussaini2022ensembles}. Formally the union of all networks $U$ is described as:

    \begin{align}
    \begin{gathered}
        U = \bigcup_{i=1}^{|N|} N_i \quad \text{with} \quad N_i \cap N_j = \emptyset \quad \forall i \neq j.
    \label{eq:trainsub}
    \end{gathered}
    \end{align}

\subsection{Efficient implementation}
\label{sec:efficientimplementation}
Our efficient implementation for VPRTempo involves training and querying a 3D weight tensor $\mathbf{T}\in \mathbb{R}^{|N|\cdot|L_i|\cdot|L_j|}$ with $|N|$ being the number of modules, $|L_i|$ the number of the neurons in the pre-synaptic layer (either $L_I$ or $L_F$), and $L_j$ being the number of neurons in the post-synaptic layer ($L_F$ or $L_O$). This setup capitalizes on parallel computing to boost efficiency and speed \cite{NEURIPS2019_9015}. During training, the weight tensor $\mathbf{T}$ is multiplied by an image tensor $\mathbf{I} \in \mathbb{R}^{|N|\cdot1\cdot|L_i|}$ that holds images associated with a particular module and their input spike rates. Single query images are fed to all modules concurrently, leveraging parallel processing to decrease the overall inference time. 
\section{Experimental Setup and Implementation}
Here we describe our implementation in \Cref{sec:implementationdetails}, the datasets used in \Cref{sec:datasets}, evaluation metrics in \Cref{sec:evalmethod}, the baseline methods that we compare and contrast our network to in \Cref{sec:baselinemethod}, and finally our strategy for optimizing system hyperparameters in \Cref{sec:paramsearch}. 
\subsection{Data processing} 
\label{sec:implementationdetails} Our network was developed and implemented in Python3 and PyTorch \cite{NEURIPS2019_9015}, converting the original BLiTNet implementation from NumPy to efficiently use multi-dimensional tensors and GPU acceleration for improved network performance. The system is trained on non-overlapping and geographically distinct places from multiple standard VPR datasets (Nordland and Oxford RobotCar \cite{olid2018single,oxfordrobot}), as commonly used in the literature \cite{hussaini2022ensembles,hussaini2022spiking,SchubertVisual, 9310189}. All reference and query input images underwent pre-processing for gamma correction, resizing, and patch normalization~\cite{6224623}. Specifically, Gamma correction was used to normalize pixels $\rho_i$ of the input images:
    \begin{align}
    \begin{gathered}
    \textit{$\rho^{norm}_i$} \ = \ 
    \begin{cases}
        \textit{$\rho^{\gamma}_i$} \quad \mbox{for} \quad 0 \geq \textit{$\rho^{\gamma}_i$} \leq 255 
        \\ 
        255 \quad \mbox{for} \quad \textit{$\rho^{\gamma}_i$} > 255,
    \end{cases}\\
    \label{eq:gamma}
    \end{gathered}
    \end{align}
where $\gamma=\frac{e^{\lambda \times 255}}{e^{\mu}}$ with $\lambda=0.5$ and $\mu=\bar{\rho_i}$. Normalized images were then resized to $W\times H=28\times 28$ pixels and patch-normalized with patch sizes $W_P \times H_P = 7\times$ 7 pixels \cite{6224623}. Neuron states $x$ are defined as floating point amplitudes in the range [0, 1], where $x = 1$ is a full spike and $x = 0$ is no spike. Initial input spikes are generated from pixel intensity values $i = [0, 255]$ from training or test images and converted to amplitudes by $x = \frac{i}{255}$.

As an abstraction of theta oscillations in the brain, the neuron state determines the timing of a spike within a timestep (Figure \ref{fig:schema}). The network hyperparameters after performing a grid search (Section \ref{sec:paramsearch}) were set as to the values in Table \ref{tab:params}. Network modules were trained to learn 1100 places with 3 modules (totalling 3300 places) from the Nordland dataset for the comparisons presented in Figure \ref{fig:trainingquery} and Table \ref{tab:results}, Figure \ref{fig:compmetrics}.%

Training each module is done simultaneously such that for each epoch, multiple unique places can be learned at the same time. After the network trained for sufficient epochs, all modules were simultaneously queried with independent network activation states to perform place matching. The network was trained for a total of 4 epochs.

\subsection{Datasets} 
\label{sec:datasets} 
We follow the suggestion by Berton et al.~\cite{Berton_2022_CVPR} to have training and reference datasets that cover the same geographic area, rather than the common VPR practice to train networks on a training set that is potentially disjoint from the deployment area (i.e.~reference dataset). Specifically, our network was evaluated on two VPR datasets, Nordland \cite{olid2018single} and Oxford RobotCar \cite{oxfordrobot} with image subsets and training performed as previously described \cite{hussaini2022ensembles}. Briefly, the Nordland dataset covers an approximately 728km traversal of a train in Norway sampled over Summer, Winter, Fall, and Spring. As is standard in the field, filtered images from the Nordland dataset removed any segments containing tunnels or speeds $<$ 15 km/hr \cite{hussaini2022ensembles,9310189,Hausler_2019}.

Images from both datasets were sub-sampled every 8 seconds, which is roughly 100 metres apart for Nordland, and 20 metres for Oxford RobotCar resulting in 3300 and 450 total places, respectively. It is important to note that while VPRSNN trained on these number of images, it reported accuracy performance on 2700 and 360 places respectively for the accuracy experiments due to the 20\% of input images used for calibration \cite{hussaini2022ensembles}; note also that our proposed VPRTempo does not require a calibration dataset. For fairness of comparison, we present accuracy measurements that have omitted the first 20\% of reference images.

\subsection{Evaluation metrics}
\label{sec:evalmethod}
When a query image $q$ is processed by the network, the matched place $\hat{p}$ is the place $i$ that is assigned to the neuron $x_i$ in the $L_O$ with the highest neuron state $x_i$:
    \begin{align}
    \begin{gathered}
        \hat{p} = \argmax_i x_i.
    \label{eq:placematch}
    \end{gathered}
    \end{align}

Network performance was measured with precision and recall \cite{SchubertVisual}. Precision is the accuracy of identified places among all the predictions made, whereas recall is the coverage of true places successfully identified from all possible true places. Recall at $N=1$ ($R@1$) measures the percent of correct matches when the network is forced to match every query image with a reference image. Recall at $N$ ($R@N$) measures if the true place is within the top $N$ matches. A match is only considered a true positive if it aligns with the exact ground truth, i.e.~no ground truth tolerance was employed.
    
    \begin{table}[t]
    \setlength{\tabcolsep}{3.9pt}
    \caption{Hyperparameters for network training}
    \centering
    \small
    \begin{tabularx}{\columnwidth}{c || c | c | c | c | c | c | c}
        Parameter & $\theta_{max}$ & $\eta^{init}_{\text{STDP}}$ & $\eta^{init}_{\text{ITP}}$ & $f_{min}, f_{max}$ & $P_{\text{exc}}$ & $P_{\text{inh}}$ & $C$ \\
        \hline
        Value & 0.5 & 0.005 & 0.15 & [0.2, 0.9] & 0.1 & 0.5 & 0.1
    \end{tabularx}
    \vspace*{-0.2cm}
    \label{tab:params}
    \end{table}

\subsection{Baseline methods}
\label{sec:baselinemethod} We evaluated our system against standard VPR methods and state-of-the-art SNN networks. Sum of absolute differences (SAD) calculates the pixel-wise absolute difference between query and reference database images, selecting a match based on the lowest sum \cite{6224623}. For SAD, images were resized to 28 x 28 pixels. NetVLAD is trained to be highly viewpoint robust, and in this case images were kept at the original resolution \cite{Arandjelovic16}. We also compare with the current state-of-the-art Generalized Contrastive Loss (GCL) \cite{leyvavallina2021gcl}. For Oxford RobotCar the dataset was resized to 320x240. Nordland and Oxford RobotCar are ``on-the-rails'' datasets with very limited viewpoint shift, which disadvantages GCL and NetVLAD. Our main comparison is VPRSNN, a recently reported SNN for performing VPR tasks based on spike rate coding by Hussaini et al.~\cite{hussaini2022ensembles}.

    \begin{table*}[!t]
    \setlength{\tabcolsep}{1.75pt}  %
    \caption{Comparison of network performance metrics}
    \centering
    \small  %
    \begin{tabularx}{\linewidth}{c | c c c c c || c c c c c}
    & \multicolumn{5}{c||}{\textbf{Nordland (3300 places)}} & \multicolumn{5}{c}{\textbf{Oxford RobotCar (450 places)}} \\
    \hline
    Method & \makecell{$R@1$ \\ (\%)} & \makecell{Train CPU\\(min)} & \makecell{Query CPU\\(Hz)} & \makecell{Train GPU\\(min)} & \makecell{Query GPU\\(Hz)} & \makecell{$R@1$\\(\%)} & \makecell{Train CPU\\(min)} & \makecell{Query CPU\\(Hz)} & \makecell{Train GPU\\(min)} & \makecell{Query GPU\\(Hz)} \\
    \hline\hline
    SAD \cite{6224623} & 48 & - & 10 & - & 17 & 37 & - & 185 & - & 126\\
    NetVLAD \cite{Arandjelovic16} & 31 & 15120 & 0.5 & 15120 & 31 & 31 & 15120 & 0.5 & 15120 & 14\\
    GCL \cite{leyvavallina2021gcl} & 32 & 360 & 2 & 360 & 77 & 33 & 360 & 0.4 & 360 & 75\\
    \hline
    VPRSNN \cite{hussaini2022ensembles} & 53 & 360 & 2 & - & - & \textbf{40} & 49 & 2 & - & - \\
    VPRTempo (ours) & \textbf{56} & \textbf{60} & \textbf{353} & \textbf{1} & \textbf{1634} & 37 & \textbf{3} & \textbf{1670} & \textbf{0.3} & \textbf{1955}\\
    \hline
    \end{tabularx}
    \label{tab:results}
    \end{table*}

\subsection{Hyperparameter search} 
\label{sec:paramsearch} To tune hyperparameters, we initially used a random search to identify the most influential parameters on match accuracy. Specifically, the hyperparameters we optimized were: firing threshold $\theta$, initial STDP learning rate $\eta_\text{STDP}^\text{init}$, initial ITP learning rate $\eta_\text{ITP}^\text{init}$, firing rate range $f_{min} \leq f_j^n \leq f_{max}$, excitatory and inhibitory connection probabilities $P_{exc}$ and $P_{inh}$, and the constant input $C$. An initial random sweep of 5,000 identified parameters for a refined grid search, specifically for $f_{min}$, $f_{max}$, $P_{exc}$, and $P_{inh}$. The resulting hyperparameters can be found in Table \ref{tab:params} and were used for both datasets. Hyperparameters were tuned on image subsets not used for any of the presented training and query performance results.

\section{Results}

In this section, we establish the network training and query speeds when run on either CPU and GPU hardware (\Cref{sec:trainingtime}). Then, we compare the performance of the network when compared to state-of-the-art and standard VPR systems (VPRSNN \cite{hussaini2022ensembles}, GCL \cite{leyvavallina2021gcl}, NetVLAD \cite{Arandjelovic16}, and SAD \cite{6224623}) (\Cref{sec:accuracy}).

\subsection{Training and querying speeds} 
\label{sec:trainingtime} 
\begin{figure}[t]
    \includegraphics[width=\columnwidth]{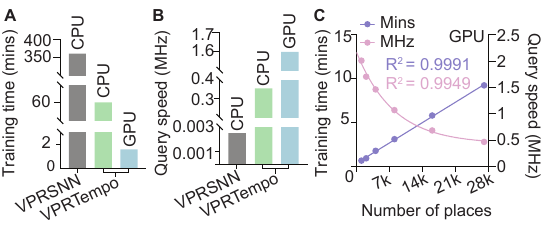}
    \caption{\textbf{A} Comparison of training times for 3300 places from the Nordland dataset our system against state-of-the-art (VPRSNN \cite{hussaini2022ensembles}). VPRSNN trained in 360 mins, VPRTempo on a CPU trained in 60 mins, with our best performance of $\approx$1 min when VPRTempo runs on a GPU. \textbf{B} Querying speed 2700 places: VPRSNN 2 Hz, VPRTempo CPU at 353 Hz, and VPRTempo GPU queried at 1634 Hz. \textbf{C} Increasing the number of places scales the training time with a time complexity of $\mathcal{O}(n)$ and inference time increases with a time complexity of $\mathcal{O}(\log\! n)$.}
        \label{fig:trainingquery}
    \vspace*{-0.3cm}
    \end{figure}

To establish network performance against a previous state-of-the-art SNN for VPR (VPRSNN), we measured the time to train each network on 3300 places from Nordland \cite{hussaini2022ensembles}. We also tested query speeds of both networks, and the effect the number of places has on both (Figure \ref{fig:trainingquery}). VPRSNN learned 3300 places in approximately 360 mins, compared to our system which required 17\% (60 mins) and 0.28\% (1 mins) of the time to train an equivalent number of images when run on CPU and GPU hardware, an Intel i7-9700K and Nvidia RTX 2080 respectively (Figure \ref{fig:trainingquery}A). Whilst our network trained the fastest on GPU hardware, CPU training was still significantly faster than VPRSNN \cite{hussaini2022ensembles}. We attribute this predominately to increased spike efficiency, shifting away from modeling biological complexity in SNN simulators \cite{10.7554/eLife.47314}.

Capable of real-time query deployment, our system inferences on CPU and GPU hardware in the order of hundreds to thousands of Hz (Figure \ref{fig:trainingquery}B), critical for resource limited compute scenarios. The time scaling of our system when training on reference datasets from 100 to 27,000 places was found to be $\mathcal{O}(n)$ for training and $\mathcal{O}(\log\! n)$ for query times, respectively (Figure \ref{fig:trainingquery}C). We next measured training and query speeds for the conventional SAD and NetVLAD and state-of-the-art GCL VPR methods, as summarized in Table \ref{tab:results}. VPRTempo trained and queried substantially faster than all of the other methods. We note that SAD has no training requirement to run and NetVLAD and GCL uses pre-trained models \cite{Arandjelovic16,leyvavallina2021gcl}. Query speed for NetVLAD and GCL was measured as the time taken for query image feature extraction \cite{Arandjelovic16,leyvavallina2021gcl}. Only CPU training or query speeds are listed for VPRSNN \cite{hussaini2022ensembles} since there is no GPU implementation of this network available.

\subsection{Network accuracy}
\label{sec:accuracy}
Now that we have established the beneficial training and inference speed, we next evaluated our system accuracy when compared to conventional and state-of-the-art VPR systems. For the Nordland dataset, our system achieved a $R@1$ of 56\% (Table \ref{tab:results}). The conventional methods SAD and NetVLAD achieved an $R@1$ of 48\% and 31\% respectively, with the state-of-the-art GCL being 32\%. Our main comparison competitor to VPRSNN achieved 53\% (Table \ref{tab:results}). 

\begin{figure}[!t]
    \includegraphics[width=\columnwidth]{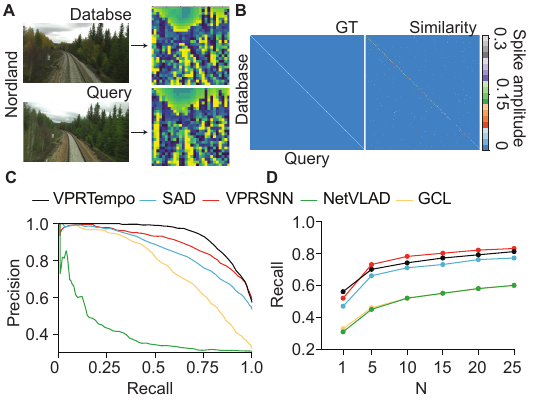}
    \caption{\textbf{A} Example database and query image from the Nordland dataset (left) that are patch-normalized for network training (right). \textbf{B} Ground truth (GT, left) and descriptor similarity matrix from testing 3300 query images (right). \textbf{C} Precision-recall curves comparing our network with sum of absolute differences (SAD \cite{6224623}), NetVLAD \cite{Arandjelovic16}, Generalized Contrastive Loss (GCL \cite{leyvavallina2021gcl}), and VPRSNN \cite{hussaini2022ensembles}. \textbf{D} Recall at N curves comparing methods as in \textbf{C}.}
    \vspace{-1em}
    \label{fig:compmetrics}
    \end{figure}

By comparison, for Oxford RobotCar dataset we achieved a $R@1$ of 37\% compared to 40\% for VPRSNN (best performer) (Table \ref{tab:results}). SAD, NetVLAD, and GCL measured at 37\%, 31\%, and 33\% respectively. In addition to this, we calculated the precision-recall curves and $R@N$ for both datasets to further validate our method, which are shown in Figure \ref{fig:compmetrics}C and D. We speculate that VPRTempo performs better in Nordland than Oxford RobotCar due to the lower viewpoint variation. As previously discussed (refer to Section \ref{sec:baselinemethod}), NetVLAD and GCL do not perform as well on static view-point datasets like Nordland and Oxford RobotCar.

\section{Conclusions and future work}

In this paper, we presented the first temporally encoded SNN to solve VPR tasks for robotic localization and navigation. We significantly improved the capacity of SNN systems to learn and query place datasets over state-of-the-art and conventional systems (SAD \cite{6224623}, NetVLAD \cite{Arandjelovic16}, GCL \cite{leyvavallina2021gcl}, VPRSNN \cite{hussaini2022ensembles}). In addition to this, we observed comparable network precision to these systems in correctly matching query to database references.

There are multiple future directions for this work and its application in robotic localization and VPR methods: 1) We are working toward translating our method onto Intel's neuromorphic processor, Intel Loihi 2 \cite{orchard2021efficient}, to deploy on energy efficient hardware. 2) For deployment on neuromorphic hardware, we are investigating the use of event streams from event-based cameras as the input to our network in order to further reduce latencies and improve energy-effciency. 3) Given our system's fast training and inference times, we will explore an ensemble fusing SNNs representing the same places for increased robustness. 4) Finally, we are exploring deployment of our network onto a robot for online and real-time learning of novel environments.

\IEEEtriggeratref{55}

\bibliographystyle{unsrt}
\bibliography{bibliography}

\end{document}